\documentclass[10pt,twocolumn,letterpaper]{article}

\usepackage{cvpr}              
\usepackage{wrapfig}
\usepackage{floatrow}
\usepackage{pifont}
\usepackage{balance}
\usepackage{multicol}
\usepackage{multirow}
\usepackage{graphicx}
\usepackage{booktabs}
\newfloatcommand{capbtabbox}{table}[][\FBwidth]

\usepackage[dvipsnames]{xcolor}

\definecolor{cvprblue}{rgb}{0.21,0.49,0.74}
\usepackage[pagebackref,breaklinks,colorlinks,citecolor=cvprblue]{hyperref}

\def\paperID{273} 
\def\confName{3DV\xspace}
\def\confYear{2025\xspace}

\title{SPAFormer: Sequential 3D Part Assembly with Transformers}

\author{
Boshen Xu\\
Renmin University of China\\
{\tt\small boshenx@ruc.edu.cn}
\and
Sipeng Zheng\\
BAAI\\
{\tt\small spzheng@baai.ac.cn}
\and
Qin Jin$^*$\\
Renmin University of China\\
{\tt\small qjin@ruc.edu.cn}
}

\begin{document}

\twocolumn[{%
\renewcommand\twocolumn[1][]{#1}%
\maketitle
\begin{center}
    \newcommand{\teaserwidth}{\textwidth}
\vspace{-0.35in}
    \centerline{\includegraphics[width=\linewidth]{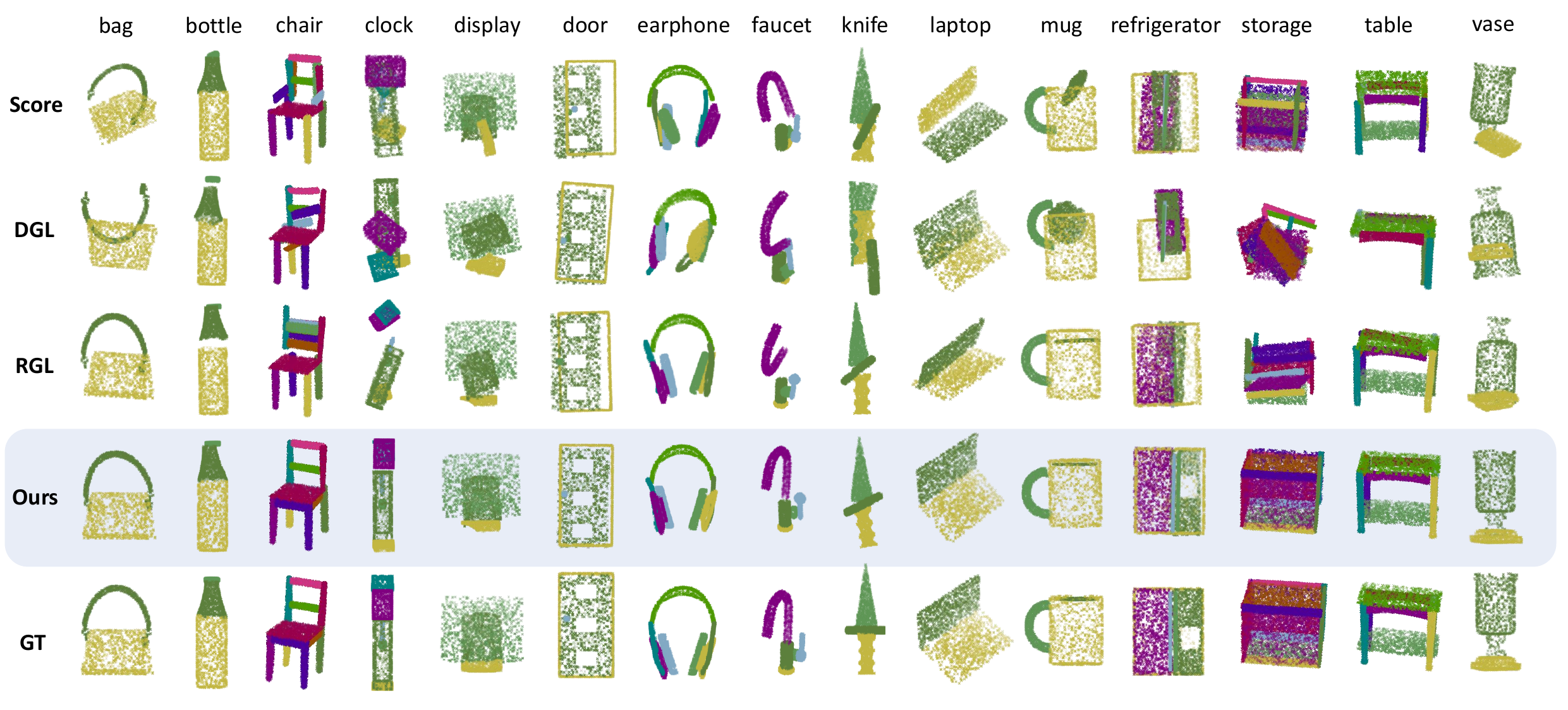}}
  \captionof{figure}{We propose SPAFormer, a novel sequence-conditioned transformer-based method to assemble objects from given parts with 3D point clouds. SPAFormer efficiently leverages the part geometry and sequence information, achieving significantly more plausible assemblies on our constructed benchmark PartNet-Assembly than other baseline methods including ScorePA~\cite{cheng2023scorepa}, DGL~\cite{HuangZhan2020DGL}, and RGL~\cite{narayan2022rgl}.}
\label{fig:teaser}
\end{center}%
}]

\renewcommand{\thefootnote}%
{\fnsymbol{footnote}}
\footnotetext[0]{*Qin Jin is the corresponding author.} 

\maketitle

\begin{abstract}
We introduce SPAFormer, an innovative model designed to overcome the combinatorial explosion challenge in the 3D Part Assembly (3D-PA) task.
This task requires accurate prediction of each part's poses in sequential steps.
As the number of parts increases, the possible assembly combinations increase exponentially, leading to a combinatorial explosion that severely hinders the efficacy of 3D-PA.
SPAFormer addresses this problem by leveraging weak constraints from assembly sequences, effectively reducing the solution space's complexity. 
Since the sequence of parts conveys construction rules similar to sentences structured through words, our model explores both parallel and autoregressive generation.
We further strengthen SPAFormer through knowledge enhancement strategies that utilize the attributes of parts and their sequence information, enabling it to capture the inherent assembly pattern and relationships among sequentially ordered parts.
We also construct a more challenging benchmark named PartNet-Assembly covering 21 varied categories to more comprehensively validate the effectiveness of SPAFormer. 
Extensive experiments demonstrate the superior generalization capabilities of SPAFormer, particularly with multi-tasking and in scenarios requiring long-horizon assembly.
Code is available at \url{https://github.com/xuboshen/SPAFormer}.
\end{abstract}

\section{Introduction}
\label{sec:intro}

\begin{figure}[t]
  \centering	\includegraphics[width=\linewidth]{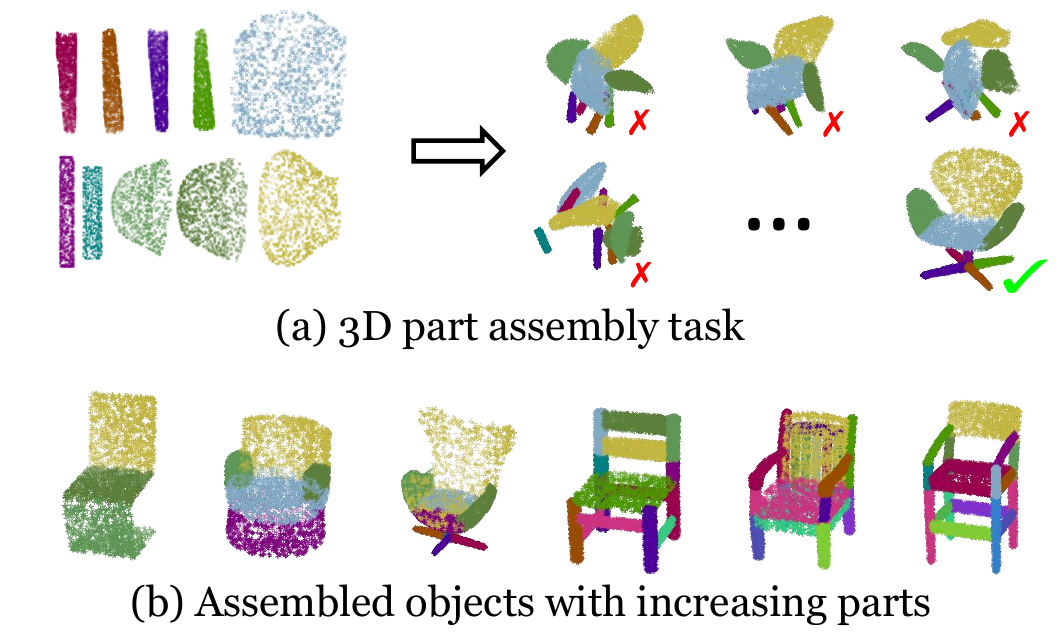}  
    \caption{Illustration of the combinatorial explosion challenge inherent in the assembly process. Specifically:
    (a) 
    For an object composed of $n$ parts, where we assume each part can occupy one of $m$ discrete positions, the potential combinations of these parts grow at an extraordinary rate, exceeding $O(m^n)$ in complexity.
    (b) the number of constituent parts increases when the target object for assembly becomes more complex.
    }
  \label{fig:explosion}
\end{figure}

There has been strong anticipation for agents to autonomously assemble parts into an intricate structural object~\cite{Seyed2022blockassembly,HuangZhan2020DGL,kataoka2023bi,wangikea,Fedor2023autoassembly}.
To achieve this goal, 3D part assembly (3D-PA) is essential~\cite{HuangZhan2020DGL}, which requires the ability of an agent to identify, reorient, and connect geometric parts in the 3D space, ultimately forming a desired object.
Such capability empowers an agent to undertake a variety of applications in real-world scenarios, ranging from assembling IKEA furniture~\cite{wangikea} to crafting designs in Computer-Aided Design (CAD) environments~\cite{Fedor2023autoassembly}.

Due to its critical importance, the 3D-PA task has attracted increasing attention from researchers.
Traditionally, many approaches~\cite{HuangZhan2020DGL,cheng2023scorepa,zhang2022ral} achieve target object assembly by relying solely on part-level information or basic structural knowledge (e.g., object category label) without additional constraints. We consider these approaches as condition-free approaches. 
However, the 3D-PA task encounters a 
substantial obstacle, known as ``combinatorial explosion'' (depicted in Figure~\ref{fig:explosion}), where the number of possible combinations rises beyond exponentially with increasing parts.
This phenomenon poses a significant challenge and makes condition-free approaches less effective in achieving accurate part assembly, particularly as the assembly sequence lengthens.

To address the combinatorial explosion challenge, several studies~\cite{narayan2022rgl,li2023gpat,li2020learningfromimg,cheng2023scorepa,ccs2023aaai,li2023category} have leveraged visual priors of the object, such as point clouds~\cite{li2023gpat} or RGB images~\cite{li2020learningfromimg, wang2022translating}, to inform the assembly process. 
These visual-conditioned methods enable the foresight of an object's final form and overall appearance to guide the assembly process.
However, the acquisition of such detailed visual priors incurs potential costs, reducing the practicality of these approaches in real-world applications.
As an alternative, recent initiatives explore a more balanced way by utilizing structural prior knowledge.
This includes the adoption of a hierarchical structure graph~\cite{ShapeScaffolder2023iccv,mo2019structurenet,gao2019sdm,jones2020shapeassembly,wang2022shape} or employing peg-hole joints on parts~\cite{li2023category} for guidance.
While this strategy decreases the need for detailed visual priors and is generally less burdensome, the reliance on structural priors still implies an additional expense compared to condition-free approaches. 
Given the limitations of previous works, we propose that leveraging the assembly sequences represents a more viable solution,
which is advantageous due to its weak constraints: assembly sequences are readily obtainable from diverse sources, including instructional manuals~\cite{wangikea, wang2022translating,Zhang2023Aligning,liu2024ikea}, the analysis of physical disassembly process~\cite{tian2022assemble, ma2023asp, sundaram2001disassembly}, or through the prediction of assembly sequences themselves~\cite{ma2023planning,tian2023asap}.
Similar to how the sequence of words in a sentence determines the meaning, the assembly order of parts conveys the rules for object construction.
Consequently, recognizing patterns within these sequences can effectively benefit 3D-PA, offering an efficient method to address combinatorial explosion.

In this work, we propose Sequential 3D Part Assembly with Transformers (SPAFormer), a novel transformer-based model to address the combinatorial explosion challenge in the 3D-PA task.
SPAFormer conditioned on the assembly sequence enhances the 3D geometric understanding through reasoning relations among object parts.
Specifically, we explore the effectiveness of potential transformer paradigms for 3D-PA, including parallel generator and autoregressive generator. 
Moreover, we employ knowledge enhancement strategies to enrich the model with part attributes and assembly sequence insights, by incorporating critical features such as part symmetry (e.g., the legs of a chair), assembly order, and relative position of the parts.
To comprehensively evaluate our model, we expand the scope of the traditional benchmark~\cite{HuangZhan2020DGL}, enriching it from the limited 3 categories (chair, table, lamp) to a wider variety of 21 categories that range from structured furniture (e.g., chairs, tables) to everyday items (e.g., earphones, clocks).
Our findings reveal that SPAFormer not only significantly outperforms existing methods that are condition-free or sequence-conditioned (\Cref{fig:teaser}), but also achieves competitive performance compared to approaches employing visual priors.

Our contributions can be summarized in three-fold:
\begin{itemize}
    \item[$\bullet$] \textbf{Innovative Framework: }
    We propose SPAFormer, a transformer-based model for 3D object assembly with sequential parts. Our model particularly leverages sequential part information and incorporates knowledge enhancement strategies to significantly improve the assembly performance.
    \item[$\bullet$] \textbf{Generalization of SPAFormer: } 
    SPAFormer shows superior generalization in object assembly from three crucial perspectives: a) category-specific, enabling it to handle various objects in the same category; b) multi-task, showcasing its versatility across diverse object categories using the same model for the first time;  and c) long-horizon, proving its ability in managing complex assembly tasks with numerous parts.
    \item[$\bullet$] \textbf{More Comprehensive Benchmark: }
    To facilitate a thorough evaluation of different models, we introduce an extensive benchmark named PartNet-Assembly, covering up to 21 object categories and providing a broad spectrum of object assembly tasks.
\end{itemize}

\section{Related Works}
\label{sec:related_works}

\noindent\textbf{Assembly-based 3D Modeling. }
The assembly-based 3D modeling involves creating structural 3D shapes by integrating information from individual parts.
Prior works~\cite{chaudhuri2011probabilistic,kalogerakis2012probabilistic,jaiswal2016assembly,funkhouser2004modeling} have explored constructing diverse 3D shapes from a given set of part candidates with precise language semantics.
With the development of generative methods, some works~\cite{wu2020pqnet,gadelha2020learning,mo2019structurenet,li2017grass,gao2019sdm} focus on leveraging generative models, such as variational autoencoders~\cite{wu2019sagnet,li2020learning}, to generate 3D shapes.
Although generative models produce aesthetic 3D shapes, they are not able to deal with assembling true objects in the real world.
Many works focus on fracture object assemblies~\cite{lu2023jigsaw,chen2022nsm,sellan2022breaking,wu2023leveraging,lamb2023fantastic,huang2006reassembling}, where the model aggregates broken pieces into an object.
However, the task of restoring broken objects is not applicable to complex furniture or toy assemblies.
Different from previous tasks, the autonomous 3D object assembly~\cite{HuangZhan2020DGL} is more beneficial for robotic assembly~\cite{kim2017parts,kataoka2023bi,Seyed2022blockassembly,zakka2020form2fit,Fedor2023autoassembly} where the parts maintain semantics like IKEA assembly.

\noindent\textbf{Autonomous 3D Object Assembly. }
In the context of 3D object assembly, the objective is to accurately predict the translation and rotation parameters of individual parts to reconstruct the object's shape.
Many studies~\cite{HuangZhan2020DGL,zhang2022ral,wang2023aaai,cheng2023scorepa} try to tackle this task by simply utilizing the part information and category-level priors.
Notably, Huang~\etal~\cite{HuangZhan2020DGL} employ graph neural networks to facilitate the understanding of interrelations between parts.
Additionally, by expanding data through the inclusion of fractured objects, Zhang~\etal~\cite{ccs2023aaai} demonstrate significant improvements in assembling chairs.
Despite these advancements, the problem of combinatorial explosion persists.
To mitigate this issue, various prior knowledge on objects has been provided as extra information, including ground-truth object image~\cite{li2020learningfromimg}, peg-hole joints on parts~\cite{li2023category} and object point clouds~\cite{li2023gpat}.
However, these constraints may be costly and less feasible in real-world applications. 
Instead, advancements in sequence-conditioned part assembly, such as RGL~\cite{narayan2022rgl}, have demonstrated the potential of using weakly constrained assembly sequences for exploring free-form assembly.
Building upon these insights, our work introduces a transformer-based framework designed to leverage the advantages offered by assembly sequences in the 3D-PA task.

\noindent\textbf{Positional Encoding in Transformers. }
Positional encoding facilitates transformers to interpret position information, as evidenced across a diverse range of research domains~\cite{xucvpr2021positional,dosovitskiy2020image,mildenhall2021nerf,li2022lepard,touvron2023llama,du2022chatglm,fang2023eva}.
Traditionally, the literature consists of two primary encodings: absolute and relative.
Absolute positional encoding~\cite{vaswani2017attention} enables transformers to differentiate words through position-dependent embeddings.
Nonetheless, it exhibits limitations in representing the order of words relative to one another~\cite{shaw2018relenc}.
To address this issue, relative positional encoding aims to better characterize the relationships between adjacent words.
A notable implementation is the RoPE~\cite{su2021roformer}, which rotates the input embedding to effectively extract relative positioning.
In the 3D-PA task, Li~\etal~\cite{li2020learningfromimg} and Narayan~\etal~\cite{narayan2022rgl} attempt to differentiate identical parts using positional encodings, but the unexpected similar features lead to their suboptimal results. 
Considering these works, we propose to combine variants of positional encodings to identify assembly patterns among sequential parts.

\section{Methodology}
\label{sec:method}

Given a set of $N$ object parts $\mathcal{P} = \{p_i\}_{i=1}^N$, the part assembly task aims to predict their poses $\{y_i\}_{i=1}^N\subset \mathrm{SE}(3)$, where $p_i$ denotes the 3D point cloud of the $i$-th part and ${\rm SE (3)}$ includes all translation and rotation transformations.
The resulting assembled 3D shape $S=\bigcup_i^N y_i(p_i)$ is formulated by transforming each part-level point cloud $p_i$ using its corresponding pose $y_i$. 
During the assembly process, the assembly chain $\mathcal{C}=\{(p_i, p_j)\mid p_i\preceq p_j,\ \forall i \leq j\}$ is imposed on the set $\mathcal{P}$, where $p_i\preceq p_j$ denotes that part $p_i$ is assembled before $p_j$. 
The goal of our work is sequential part assembly, which predicts the poses $\{y_i\}_{i=1}^N$ of $\mathcal{P}$ guided by the predefined assembly chain $\mathcal{C}$. 

\begin{figure*}[t]
  \centering
  \includegraphics[width=\textwidth]{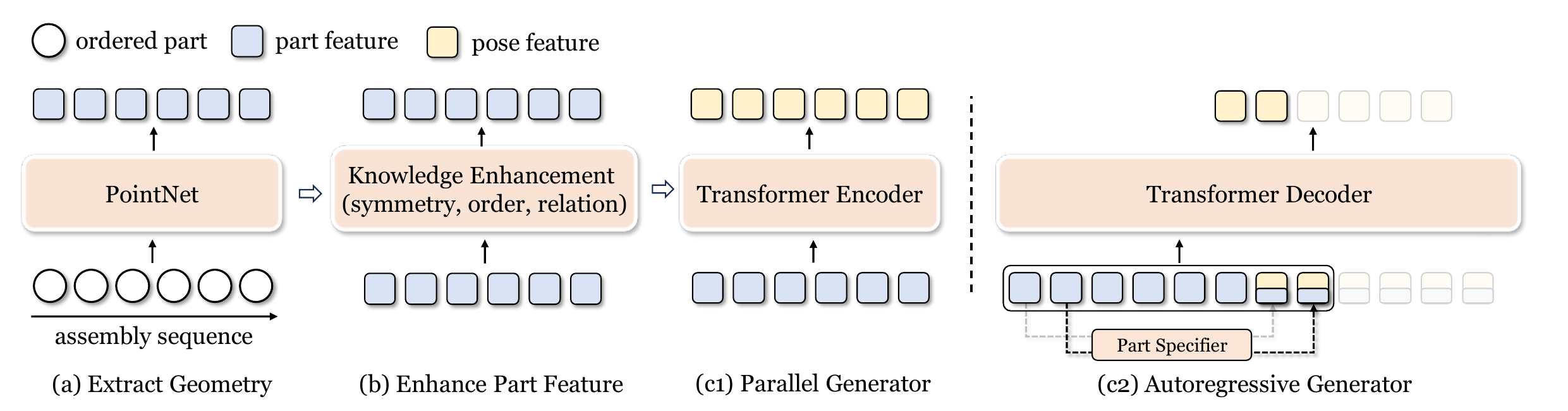}  
    \caption{Illustration of overall end-to-end framework of SPAFormer. (a) The shared 3D backbone extracts the geometry feature of individual parts, followed by (b) knowledge enhancement of part features, which incorporates symmetry, order, and relation information into part features through positional encodings, then generates poses by either (c1) parallel generator, which generates poses of all parts at once, or (c2) autoregressive generator, which decodes poses of parts according to assembly sequences step by step.}
  \label{fig:model_arch}
    \vspace{-8pt}
\end{figure*}

The overall framework of our proposed SPAFormer is illustrated in Figure~\ref{fig:model_arch}.
In the following sections, we first introduce our knowledge enhancement strategies to effectively incorporate part attributes and relation information in Section~\ref{subsec:pos_enc}.
Then we discuss generator variants for better modeling part relationship in Section~\ref{subsec:transformer}.
These strategies include order encoding (OEnc) and relation encoding (REnc) which aim to capture assembly sequence patterns, and symmetry encoding (SEnc) which focuses on enriching part features with symmetry information.
In Section~\ref{subsec:train_inf_pipeline}, we introduce our training process.

\subsection{Knowledge Enhancement for Parts}
\label{subsec:pos_enc}
Given the point clouds of object parts constrained by assembly sequence $\mathcal{C}$, a shared 3D encoder~\cite{qi2017pointnet} is firstly used to extract their geometric representation $\mathcal{V}_0=\{v_i\}_{i=1}^N$, which contains the shape knowledge of each part. 

Due to the data deficiency, which can lead to inferior results when using the data-hungry transformers, we employ several strategies to enhance the model's understanding of parts.
These strategies include order encoding (OEnc) $h(\cdot)$, relation encoding (REnc) $f_{\{q,k\}}$ to enhance the model's understanding of neighbor parts, and symmetry encoding (SEnc) $g(\cdot)$ for modeling parts attributes.

\noindent\textbf{Order Encoding.}
To leverage the assembly pattern in part sequence, order encoding reveals the order information to transformers:  $h(v_i, i)=v_i\oplus e_N(i)$, where $\oplus$ denotes feature concatenation and $e_N(i)$ is a one-hot embedding with the $i$-th position being 1 in a vector of length $N$. 
This encoding integrates the $i$-th order information into the part feature $v_i$ before passing the part feature to the transformers.

\noindent\textbf{Relation Encoding.}
We introduce relation encoding to capture the relative positions of assembling parts such as the bottom-to-top assembly order in IKEA furniture. 
Relation encoding is implemented using rotary embedding~\cite{su2021roformer} to enhance the attention mechanism, which rotates the representations of query-key pairs in self-attention. 
We leverage its long-term decay property on attention weights to encode relative relations between parts:

\begin{equation}
    f_{\{q,k\}}(v_i, i)=R_{\Theta, i}^d v_i
\end{equation}
where $R_{\Theta, i}^d=\operatorname{diag}(A_1, \dots, A_{d/2})$, $A_i=\begin{bmatrix}
\cos(m\theta_i) & -\sin(m\theta_i)\\
\sin(m\theta_i) & \cos(m\theta_i)
\end{bmatrix}
$ is a rotation matrix, and $\operatorname{diag}(\cdot)$ denotes diagonalizing block matrices.
The relation encoding is applied to queries and keys to capture the relative order information:
\begin{equation}
    q_i\cdot k_j = (R_{\Theta, i}^d v_i)\cdot (R_{\Theta, j}^d v_j)=v_i\cdot R_{\Theta, i-j}^d v_j
\end{equation}
where $\cdot$ denotes dot product, and the attention weights decay as the relative position between query and key increase. 
We refer to Su~\etal~\cite{su2021roformer} for details.

\noindent\textbf{Symmetry Encoding.}
Symmetrical relations, such as translational (e.g., identical and parallel arms of a chair), reflective (e.g., mirror-symmetrical arms of a chair), and rotational symmetry (e.g., identical but not parallel legs of a chair), are proven beneficial for creating structured objects in generative methods~\cite{mo2019structurenet}.
To incorporate symmetrical relations into part features, we use principal component axes via PCA to acquire 3D bounding boxes and calculate bounding box sizes, which are invariant to input poses.
The parts are organized based on identical bounding box sizes, resulting in the partition $\mathcal{P}=\{\mathcal{P}_1, \cdots, \mathcal{P}_M\}$, where each part $p_i$ is associated with a unique group $s_i\in \{1,\dots,M\}$.
Symmetry encoding explicitly models symmetrical information into part features: $g(v_i)=v_i \oplus e_N(s_i)$.

Consequently, symmetrical parts within the groups share similar features, producing distinguishable predictions between different groups.
To prevent collapsed prediction within a group, bipartite matching is conducted inside each group $\mathcal{P}_i=\{p_{i}^j\}_{j=1}^{n_i}$ between the transformed parts $\{y_{i}^j (p_{i}^j)\}_{j=1}^{n_i}$ and the ground truths $\{p_{i}^{j*}\}_{j=1}^{n_i}$, where the asterisk symbols ($*$) represents the respective ground truths in all equations.
While previous works~\cite{li2020learningfromimg, HuangZhan2020DGL, narayan2022rgl} assign distinct encodings to symmetrical parts inside a group $\mathcal{P}_i$ to handle the ambiguity prediction of symmetrical parts, we suggest distinguishing different groups and assemble symmetrical parts by bipartite matching.

\subsection{Generator Variants for Part Assembly}
\label{subsec:transformer}
Then we discuss different generator variants to predict the 3D poses of object parts. 

\noindent\textbf{Parallel Generator. }
As illustrated in Figure~\ref{fig:model_arch} (c1), this is the most straightforward way adopted by 3D-PA approaches~\cite{HuangZhan2020DGL,ccs2023aaai}.
The parallel generator simultaneously outputs the 3D poses of the sequential parts in parallel, which can be denoted as:

\begin{equation}
\small
    \{y_{i}\}_{i=1}^N = \mathcal{G}(\mathcal{F}(\{v_i\}_{i=1}^N))
\end{equation}
where $\mathcal{F}$ denotes the transformer encoder~\cite{vaswani2017attention} with $L$ layers, which aims to build relationships among parts using self-attention, and $\mathcal{G}$ is a shared MLP that maps part representations to poses.

\noindent\textbf{Autoregressive Generator. }
Instead of parallel generation, another alternative is to autoregressively decode the 3D poses of object parts following the assembly sequence.

To provide a global context of the potential assembled object, all part features are prepended as prompts before decoding the poses. 
At each step $i$, the model generates a pose feature $z_i$, concatenates it with the next part feature $v_{i+1}$, and feeds it to the transformer model as input recursively.
To decide the next part and maintain consistency in dimension, a part specifier indexes the upcoming part following the assembly sequence and projects the part feature for concatenation with the last pose feature, as depicted in~\cref{fig:model_arch} (c2). 
We use two linear projectors $\phi_1$ and $\phi_2$ to downsample the part $v$ and pose $z$ features from $\mathbb{R}^{d}$ to $\mathbb{R}^{\frac{d}{2}}$, respectively, where $d$ is the hidden dimension of the transformers.
As a result, the generation progress is formulated as:
\begin{equation}
\small
    y_{i} = \mathcal{G}(\mathcal{F}(\{v_k\}_{k=1}^N; \{(\phi_{1}(z_{j})\oplus\phi_{2}(v_{j+1})\}_{j=0}^{i-1}))
\end{equation}

\subsection{Training Process}
\label{subsec:train_inf_pipeline}

\noindent\textbf{Training Loss.}
Supervision of both individual parts and the entire shape is applied during training.
The part pose $y_i$ is represented by translation $t_i\in \mathbb{R}^3$ and rotation $r_i\in \mathbb{R}^4$ in the form of quaternion. The translation is supervised via a MSE loss:
\begin{equation}
    \mathcal{L}_t = \sum\limits_{i=1}^N \|t_i-t_i^*\|_2^2
\end{equation}
The distance between two sets of point clouds $\mathcal{X}_1, \mathcal{X}_2$ can be measured by Chamfer distance~\cite{fan2017point}:
\begin{equation}
    \mathcal{CD}(\mathcal{X}_1, \mathcal{X}_2) = \sum\limits_{i,j\in\{1,2\}} \sum\limits_{x_i\in \mathcal{X}_i} \min_{x_{j}\in \mathcal{X}_{j}} \|x_{i}-x_{j}\|_2^2
\end{equation}

Since rotation is not unique, the rotation is supervised via Chamfer Distance on the rotated part point cloud instead of directly applying an MSE loss on $r_i$:
\begin{equation}
    \mathcal{L}_r = \sum\limits_{i=1}^N(\mathcal{CD}(y_i(p_i), y_i^*(p_i)))
\end{equation}
In order to better optimize global shapes, we also apply Chamfer Distance on the complete shape:
\begin{equation}
    \mathcal{L}_s = \mathcal{CD}(S, S^*)
\end{equation}
The final objective for 3D part assembly is calculated as:
\begin{equation}
    \mathcal{L} = \lambda_t\mathcal{L}_t + \lambda_r\mathcal{L}_r + \lambda_s\mathcal{L}_s
\end{equation}
where $\lambda_t$, $\lambda_r$ and $\lambda_s$ are hyper-parameters.

\section{Experiments}
\label{sec:exp}
\subsection{Experiment Setup}
\label{subsec:setup}
\noindent\textbf{Dataset Setup. }
We notice that prior studies only consider assembling merely three object categories (chair, table, lamp), which is a far cry from diverse objects in reality.
To this end, we expand the available objects to include 21 categories from PartNet~\cite{mo2019partnet}, thereby introducing a more diversified set of objects to verify the SPAFormer's generalization. 
We refer to this enhanced benchmark as PartNet-Assembly,
and it is split into two subsets: (1) \textit{furniture object}, containing three structurally complex categories (chair, table, and storage) that offer sufficient and diverse samples, and (2) \textit{daily object}, containing the remaining 18 categories that represent commonly used everyday items (e.g. earphone).
In total, PartNet-Assembly contains 22,873 shapes from 21 object categories with 173,765 distinct parts.
The dataset is partitioned into training (70\%), validation (10\%), and testing (20\%) sets, following the distribution protocol proposed by Mo~\etal~\cite{mo2019partnet}.

\noindent\textbf{Implementation Details.}
We utilize the Adam optimizer initialized with a learning rate of 1.5e-4, and employ an exponential decay strategy, reducing the learning rate by a factor of 0.8 every 80 epochs. 
The training process is conducted on 2 GPUs, employing a batch size of 128, over a total of 800 epochs. 
The optimal epoch is selected through performance assessment using Part Accuracy (PA) metrics on the unseen validation split.
The transformer architecture is configured with a hidden dimension size of 512 across 6 layers.
Following Huang~\etal~\cite{HuangZhan2020DGL}, the hyperparameters are set as follows: $\lambda_t=1$, $\lambda_r=10$, $\lambda_s=1$.
Furthermore, all target shapes and input parts within PartNet-Assembly are normalized to the canonical space using PCA, which eliminates the impact of initial poses.

\noindent\textbf{Evaluation Metrics.}
In this study, we evaluate the quality of the assembled shapes using four key metrics:

\begin{itemize}
    \item[$\bullet$] Shape Chamfer Distance ({\rm SCD})~\cite{li2020learningfromimg},
    measures the overall distance between the predicted and ground-truth shapes using the equation below, with all values scaled by $10^{3}$ for simplicity:
    \begin{equation}
        \text{SCD} = \mathcal{CD}(\cup_{i=1}^N y_i(x_i), \cup_{i=1}^N y_i^*(x_i))
    \end{equation}
    
    \item[$\bullet$] Part Accuracy ({\rm PA})~\cite{li2020learningfromimg}, assesses the correctness of part placement. A part is considered accurately placed if its error distance is below a specified threshold $\epsilon=0.01$:
    \begin{equation}
        \text{PA} = \mathbb{I} \left[ \mathcal{CD}(y_i(x_i), y_i^*(x_i)) < \epsilon \right]
    \end{equation}
    
    \item[$\bullet$] Connectivity Accuracy ({\rm CA})~\cite{HuangZhan2020DGL}, measures the quality of connections between adjacent part pairs in the assembled shape. With $\tau=0.01$ as the threshold, and $C=\{(c_{ij}, c_{ji})\}$ representing the set of all connected point pairs between adjacent parts $(p_i, p_j)$:
     \begin{equation}
        \text{CA} = \frac{1}{|C|} \sum_{(c_{ij}, c_{ji})\in C} \mathbb{I}\left[ (y_i(c_{ij})-y_j(c_{ji}))^2 < \tau \right]
    \end{equation}
    
    \item[$\bullet$] Success Rate ({\rm SR})~\cite{li2023gpat}, is defined as 1 for an object if the part accuracies of all components are equal to 1.
\end{itemize}

\subsection{Ablation Study}

\noindent\textbf{1. Effect of different generator variants for assembly. }

In Table~\ref{tab:model_arch}, we compare the performance of different variants outlined in Section~\ref{subsec:transformer}  on the table assembly task.
Both variants implement identical knowledge enhancement strategies.
The results indicate that the parallel generator (denoted as "para") outperforms the autoregressive generator (denoted as "auto") across a majority of the 3D-PA metrics, with the exception of CA.
Specifically, the "para" generator excels in SCD, PA, and SR, highlighting its proficiency in assembling objects that are globally coherent.
Conversely, the "auto" generator brings gains in CA, indicating its ability to ensure better connectivity among adjacent parts of the objects.
Due to the ease of training, we adopt the ``para'' generator in our following experiments.

\begin{table}[t]
\centering
\caption{Comparison of different generator variants on table assembly, where ``para'' and ``auto'' denote the parallel and autoregressive generator, respectively.}
\scalebox{0.8}{
\begin{tabular}{c|ccccc}
\toprule
Generator & SCD↓ & PA↑ & CA↑ & SR↑ \\
\midrule
para & 3.80 & 64.38 & 57.60 & 33.50\\ 
auto & 4.14 & 62.18 & 59.41 & 31.92\\
\bottomrule
\end{tabular}
}
\label{tab:model_arch}
\end{table}

\begin{table}[t]
\centering
\caption{Comparison of different assembly sequence patterns on chair assembly.}
\scalebox{0.8}{
\begin{tabular}{c|cccc}
\toprule
Pattern & SCD↓ & PA↑ & CA↑ & SR↑ \\
\midrule
diagonal & 6.74 & 55.88 & 36.39 & 16.40 \\ 
top-to-bottom & 9.36 & 45.29 & 25.45 & 10.30 \\ 
bottom-to-top & 9.07 & 45.72 & 27.21 & 11.24 \\ 
descending size & 10.88 & 38.09 & 21.91 & 9.65 \\ 
random & 14.38 & 29.40 & 16.69 & 5.99\\
\bottomrule
\end{tabular}
}
\label{tab:train_seq}
\end{table}

\noindent\textbf{2. Effect of different assembly sequence patterns.}

We carry out ablations in Table~\ref{tab:train_seq} to evaluate the effectiveness of different assembly sequences:
a) diagonal, which involves assembling an object from one corner diagonally to the opposite corner (e.g., assembly a chair from a front leg to the back);
b) top-to-bottom, where objects are assembled from the top down;
c) bottom-to-top, which is the reverse process of top-to-bottom, where assembly starts from the bottom;
d) descending size, which is determined by the sizes of object components, calculated via principal component analysis (PCA), starting with the largest.
The outcomes reveal that the diagonal pattern outperforms the others.
This superior performance can be attributed to its ability to sequentially assemble symmetric components (such as the four legs of a chair), a feat that is more efficiently accomplished than other patterns like top-to-bottom.
Based on these findings, we adopt the diagonal assembly pattern as our default strategy in subsequent experiments.

\begin{table}[t]
\centering
\caption{Comparison of different knowledge enhancement strategies on chair assembly, where OEnc, REnc, and SEnc denote order, relation, and symmetry encoding, respectively.}
\setlength{\tabcolsep}{5pt}
\scalebox{0.8}{
\begin{tabular}{c|ccc|cccc}
\toprule
 & OEnc & REnc & SEnc  & SCD↓ & PA↑ & CA↑ & SR↑ \\
\midrule
1 & \ding{55}&\ding{55}&\ding{55}   & 14.38 & 29.40 & 16.69 & 5.99\\
2 & \ding{55} &\ding{51}&\ding{55}  & 14.71 & 28.45 & 16.38 & 5.34\\
3 & \ding{55} &\ding{55}&\ding{51}  & 13.35 & 37.48 & 21.41 & 7.68\\
4 & \ding{51} &\ding{55}&\ding{55}  & 8.30 & 47.35 & 29.57 & 12.08\\
5 & \ding{51}&\ding{51} &\ding{55}  & 8.04 & 48.51 & 30.61 & 12.18\\
6 & \ding{51}&\ding{55} &\ding{51}  & 7.91 & 54.17 & 35.31 & 14.05\\
7&\ding{51} & \ding{51}&\ding{51}  & \textbf{6.74} & \textbf{55.88} & \textbf{36.39} & \textbf{16.40}\\
\bottomrule
\end{tabular}
}
\label{tab:abl_on_encoding}
\end{table}

\noindent\textbf{3. Effect of knowledge enhancement strategies.}

\begin{figure}[t]
\caption{Visualizations of assembly results when enhancing knowledge by adding new encoding patterns in a stepwise way.}
    \centering
    \includegraphics[width=0.9\linewidth]{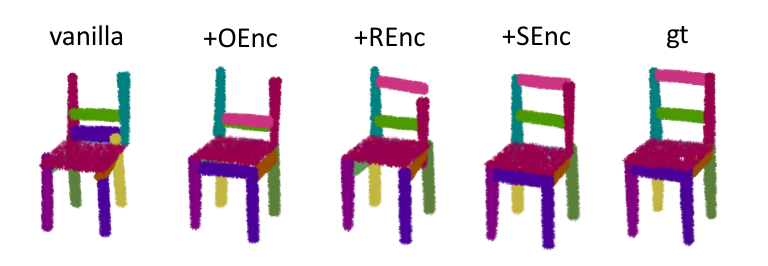}
\label{fig:quanlitative_ablation}
\end{figure}

Table~\ref{tab:abl_on_encoding} reveals the impact of various encoding strategies on 3D part assembly.
As can be seen, each strategy can bring improvements.
For example, employing OEnc and SEnc augment the assembly outcomes by $+8.08\%$ and $+17.95\%$ on PA, by comparing Row 3-4 with Row 1. 
This demonstrates the value of incorporating order and symmetry information into the model, which significantly benefit the assembly process.
However, while REnc alone does not notably advance assembly capabilities in Row 2, its combination with other encodings acts as a pivotal enhancer, akin to a catalyst.
For instance, the comparison between Row 6 and Row 7 reveals that the collaborative effect of REnc with other encodings improves SR by $+2.35\%$ and PA by $+1.71\%$.
Qualitative examples are provided in Figure~\ref{fig:quanlitative_ablation} to present the gains of integrating each encoding strategy.

\noindent\textbf{4. Effect of varied assembly sequence length.}

\begin{figure}[t]
  \centering
    \includegraphics[width=0.9\linewidth]{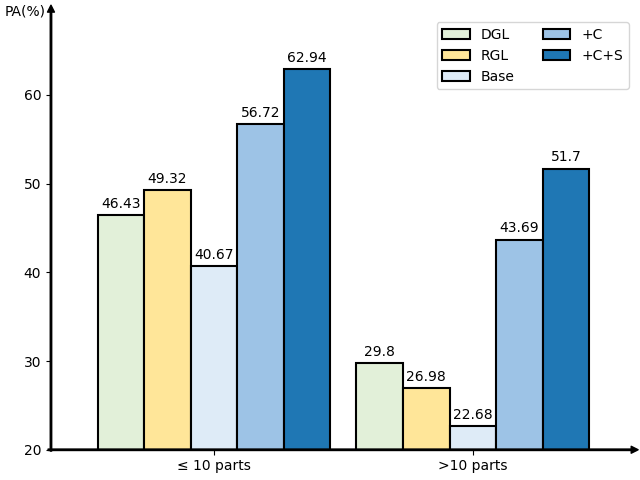}
  \caption{Comparison of varied assembly length. Our model presents notable improvements particularly in long-horizon assembly ($>$10 parts) when the model is enhanced by incorporating OEnc and REnc (+C), as well as SEnc (+S).}
  \label{fig:assembly_length}
\vspace{-0.3cm}
\end{figure}
\begin{table*}
\centering
\caption{Comparison with previous SoTA works on the furniture assembly task.}
\setlength{\tabcolsep}{2.2pt}
\scalebox{0.9}{
\begin{tabular}{lc|ccc|ccc|ccc|ccc}
\toprule
&&\multicolumn{3}{c|}{SCD↓}& 
\multicolumn{3}{c|}{PA↑}&
\multicolumn{3}{c|}{CA↑}&
\multicolumn{3}{c}{SR↑}
\\
\midrule
Method & Seq & Chair & Table & Storage & Chair & Table & Storage &
Chair & Table & Storage & Chair & Table & Storage \\
\midrule
DGL~\cite{HuangZhan2020DGL} & \ding{55} & 9.1 & 5.0 & 12.1 & 39.00 & 49.51 & 12.39 &
23.87 & 39.96 & 17.23 & 8.44 & 20.35 & 0.00\\
IET~\cite{zhang2022ral} & \ding{55} & 12.3 & 5.6 & 6.8 & 39.30 & 49.21 & 29.31 &
17.83 & 35.82 & 22.51 & 6.18 & 21.34 & 1.07 \\
Score-PA~\cite{cheng2023scorepa} & \ding{55} & 7.4 & 4.5 & 6.9 & 42.11 & 51.55 & 27.68 &
34.27 & 32.79 & 21.82 & 8.32 & 11.23 & 0.35 \\
CCS~\cite{ccs2023aaai} & \ding{55} & \underline{7.0} & - & - & 53.59 & - & - &
\textbf{38.97} & - & - & - & - & - \\
\midrule

Complement~\cite{sung2017complementme} & \ding{51} & 43.4 & 16.5 & 29.3 & 9.85 & 12.90 & 8.61 &
19.39 & 22.75 & 31.17 & 0.00 & 5.94 & 3.57 \\

LSTM~\cite{wu2020pqnet} & \ding{51} & 18.7 & 8.0 & 11.9 & 11.21 & 25.63 & 8.50 &
12.30 & 29.66 & 27.14 & 0.46 & 8.79 & 0.35\\

RGL~\cite{narayan2022rgl} &\ding{51} & 10.9 & 5.5 & 6.4 & 42.16 & 59.99 & 48.02 &
24.85 & 49.89 & 38.45 & 7.12 & 26.43 & 3.92 \\

\midrule
Ours-specific & \ding{51} & \textbf{6.7} &\textbf{ 3.8} & \underline{4.5} & \textbf{55.88} & \underline{64.38} &\underline{ 56.11} &
36.39  & \underline{57.60} & \underline{49.98}  & \textbf{16.40 }& 33.50 & 7.85\\
Ours-multitask & \ding{51} & \underline{7.0} &\underline{4.4} & \textbf{3.8} & \underline{55.84} &\textbf{64.44} & \textbf{67.14} &
\underline{38.75}  & \textbf{61.04} & \textbf{65.20}  & \underline{13.96}& \textbf{33.57} & \textbf{15.71}\\
\bottomrule
\end{tabular}}
\label{tab:specialized}
\end{table*}

\begin{table}[t]
\centering
\caption{Comparison of daily object assembly on 18 categories.}
\scalebox{0.9}{
\begin{tabular}{c|cccc}
\toprule
Method & SCD↓ & PA↑ & CA↑ & SR↑ \\
\midrule
Score-PA~\cite{cheng2023scorepa} & \textbf{6.38} & 32.39 & 32.16 & 15.05 \\
DGL~\cite{HuangZhan2020DGL} & 6.57 & 36.55 & 40.01 & 18.73 \\ 
RGL~\cite{narayan2022rgl} & 10.93 & 51.33 & 50.52 & 32.00 \\ 
\midrule
Ours & 8.61 & \textbf{54.32} & \textbf{51.60} & \textbf{37.12} \\ 
\bottomrule
\end{tabular}
}
\label{tab:daily_obj}
\end{table}

We conduct an analysis of varied assembly lengths for the identification of assembly patterns and symmetry information in Figure~\ref{fig:assembly_length}.
It is generally accepted that assembling objects composed of more parts presents a higher degree of difficulty.
To this end, we categorize the testing set into two sets based on the number of parts: short-horizon assembly, which consists of fewer than or equal to 10 parts, and long-horizon assembly, encompassing those with more than 10 parts.
For example, a chair usually comprises fewer than 10 parts including four legs, a back, and a seat. 
Compared with the vanilla attention baseline, our model that captures sequence information (denoted as ``+C'') not only exhibits a significant performance improvement of $+16.05\%$ in short-horizon assembly but also shows an impressive gain of $+21.01\%$ in the more complex long-horizon assembly.
Further analysis with SEnc (marked as ``+S'') demonstrates the use of SEnc as a versatile technique that uniformly enhances PA performance across different assembly lengths.
Such results indicate that our model is able to effectively capture complex structural information, marking a promising avenue for tackling challenging long-horizon tasks.
In contrast, existing approaches~\cite{narayan2022rgl, HuangZhan2020DGL} struggle with long-horizon assembly, with a decrease in PA exceeding $-21.9\%$.

\noindent\textbf{5. Effect of multi-task joint training.}

\begin{table}[t]
\centering
\caption{Comparison between category-specific and universal implementations of SPAFormer, evaluated on 21 object categories.}
\setlength{\tabcolsep}{4pt}
\scalebox{0.9}{
\begin{tabular}{c|cccc}
\toprule
 & SCD↓ & PA↑ & CA↑ & SR↑ \\
\midrule
ours-specific & 6.54 & 57.79 & 49.95 & 28.80 \\ 
ours-multitask & 6.37 & 59.78 & 53.74 & 29.40 \\ 

\bottomrule
\end{tabular}
}
\label{tab:multi_task}
\end{table}

An ideal assembly agent should be proficient in assembling a wide array of objects through a unified model.
However, the exploration of such a universal assembly methodology remains underrepresented in existing works, and the category-specific approach is still widely adopted.
To address this gap, our work delves deeper into the feasibility of a general-purpose assembly approach.
In addition to vanilla category-specific SPAFormer (denoted as ``ours-specific'') where each model is tailored to a specific category, we develop a multi-task model (denoted as ``ours-multitask'') that is trained across all accessible object categories.
We compare the two implementations in Table~\ref{tab:multi_task}.

The results reveal that the multi-task model excels in assembling objects from categories with both abundant (e.g., tables, storage furniture) and scarce (e.g., mugs with 120 samples, bowls with 130 samples) training data (Details can be found in supplementary). 
To be specific, the multi-task model presents a modest enhancement of $+1.99\%$ on PA, +3.79\% on CA, and +0.6\% on SR, highlighting the viability of a general-purpose approach to part assembly. 
Given these positive outcomes, we intend to 
further explore towards this direction, focusing on identifying which categories most benefit from shared assembly knowledge and controlling the generated categories in future works.

\subsection{Comparison with State of the Arts}

\label{subsec:cat_pa}
\noindent\textbf{Comparison of Furniture Assembly. }
Assembling furniture is challenging due to the complex and varied shapes of structured objects, which troubles many assembly methods.
As shown in Table~\ref{tab:specialized}, our model significantly outperforms both previous score-based and graph-based models in the assembly of structured furniture objects across almost all evaluated metrics.
For instance, when assembling unseen tables, our category-specific approach (denoted as ``Ours-specific'') remarkably exceeds the performance of the leading prior model, RGL~\cite{narayan2022rgl}, achieving improvements of $+4.39\%$ on PA, $+7.71\%$ on CA, and $+7.07\%$ on SR.
These results demonstrate the efficacy of our proposed SPAFormer in handling furniture assembly tasks.

Additionally, we delve into a comparison of various conditional assembly techniques with a focus on the chair assembly task due to its structural property, as presented in Table~\ref{tab:comp_constraints}.
To ensure fair comparisons, we conduct experiments on the dataset adopted by Li~\etal~\cite{li2020learningfromimg} and evaluate these models on PA and SR, which are fairly compared metrics across different constraints. 
Despite being compared against methods that utilize ground truth inputs such as images~\cite{li2020learningfromimg} or 3D point clouds~\cite{li2023gpat}, our model demonstrates commendable performance.
This achievement highlights the distinct advantage of our sequence-conditioned assembly approach, which effectively generalizes to novel structures solely based on the relationships between parts, without relying on external visual information.

\begin{table}[t]
\centering
\caption{Comparison of conditional assembly on chair assembly.}
\scalebox{0.9}{
\begin{tabular}{c|c|cc}
\toprule
Method & Constraints & PA↑ & SR↑ \\
\midrule
Score-PA~\cite{cheng2023scorepa}& condition-free & 42.11 & 8.32 \\
Img-PA~\cite{li2020learningfromimg} & image gt& 49.10 & - \\
GPAT~\cite{li2023gpat} & point cloud gt& \textbf{57.7}  & 19.30 \\ 
\midrule
Ours & sequence & 57.29 & \textbf{21.97} \\ 
\bottomrule
\end{tabular}
}
\label{tab:comp_constraints}
\end{table}

\noindent\textbf{Comparison of Daily Object Assembly.}
To assess our model's performance in broader scenarios, we conduct experiments on the collection of 18 everyday objects (e.g. earphone, bottle), showcasing its capability to generalize across a wider range of items.
As illustrated in Table~\ref{tab:daily_obj}, our SPAFormer demonstrates a substantial advantage over condition-free methods, achieving an improvement of at least $+18.39\%$ on SR, and surpassing another sequence-conditioned RGL method by $5.12\%$ on SR.
Despite condition-free approaches~\cite{cheng2023scorepa, HuangZhan2020DGL} performs better on SCD, they significantly underperform across other metrics.
We attribute this to the optimization collapse, where the value of SCD loss is low but the assembled shapes are less reasonable, as shown in Row 1-2 in Figure~\ref{fig:teaser}.

\subsection{Qualitative Analysis}

\begin{figure}[t]
\caption{Visualizations of failed object assemblies, including rare shapes, long-horizon assembly, and wrong orientation.} 
        \centering
        \includegraphics[width=0.9\linewidth]{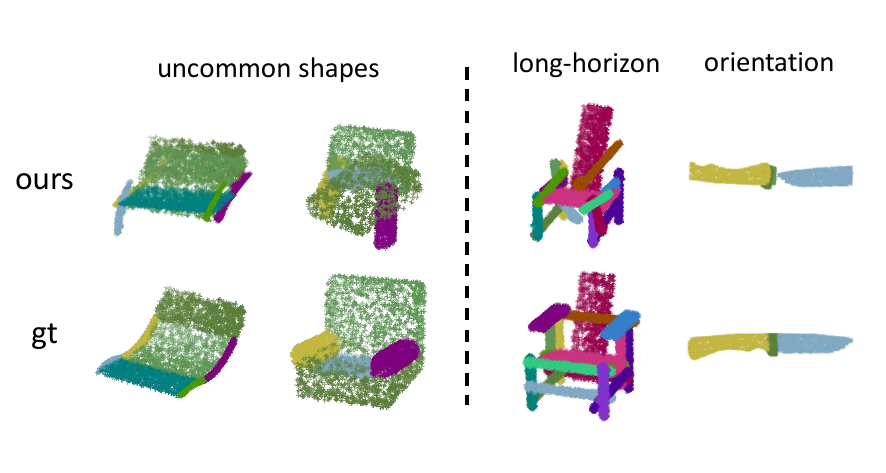}
        \label{fig:failure_cases}
\end{figure}

\noindent\textbf{Failure cases.}
In exploring the limitations of our model, Figure~\ref{fig:failure_cases} highlights two primary factors leading to assembly failures: out-of-distribution (OOD) shapes and our model's inherent weakness.
The left two columns in Figure~\ref{fig:failure_cases} depict OOD shapes that are rarely encountered in reality, such as recliner chair and sofa without chair legs.
The examples on the right side of Figure~\ref{fig:failure_cases} shed light on inherent weaknesses in our model.
Despite making significant strides in addressing long-horizon assembly challenges as evidenced in Figure~\ref{fig:assembly_length}, our model exhibits limitations in assembling complex objects, such as chairs composed of up to 17 parts.
Furthermore, our model encounters difficulties in accurately recognizing part orientations, primarily due to its limited ability to detect the junctions between parts.
To address these issues, labeling parts with peg-hole joints~\cite{li2023category} and reframing the problem as bipartite matching is a costly alternative.

\noindent\textbf{Real-world experiment.} Additionally, we conduct a real-world experiment, where the object scan is sourced from the Redwood dataset~\cite{redwood}, to successfully assemble a table by our model, suggesting potential generalization to real-world scenarios where the part point clouds are imperfect.
\begin{figure}[t]
  \centering
  \includegraphics[width=\linewidth]{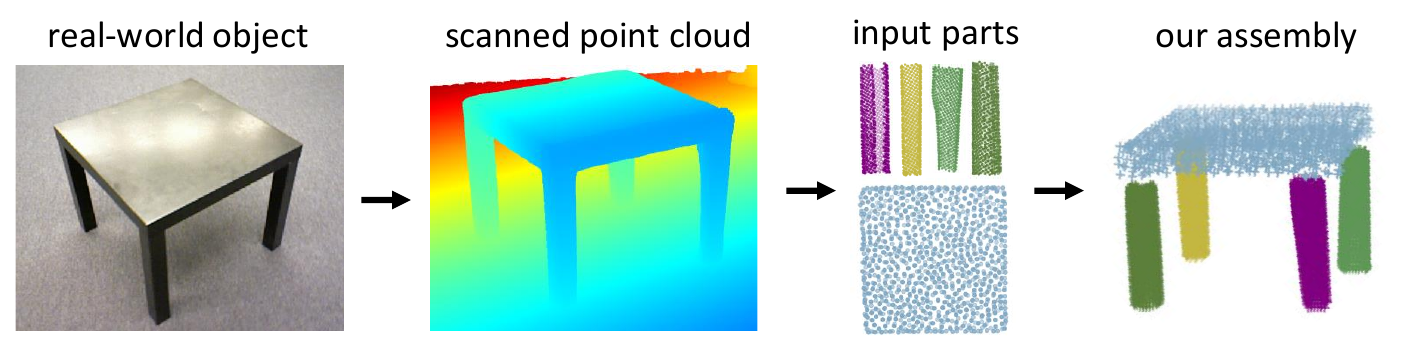}
   \caption{Real-world experiment on table assembly.}
   \label{fig:real_world}
\end{figure}

\section{Conclusion}
In this study, we introduce a novel model SPAFormer, which is designed to tackle the issue of combinatorial explosion in 3D part assembly tasks.
Building upon insights from prior research, SPAFormer employs assembly sequences to effectively narrow down the extensive solution space.
We explore different assembly generator variants and incorporate multiple knowledge enhancement strategies for capturing the inherent assembly pattern.
To facilitate comprehensive evaluation, we extend the existing benchmark to PartNet-Assembly, which spans 21 diverse categories including both furniture and daily objects. 
Experimental analyses are conducted to validate components of SPAFormer and the enhanced performance over existing methods.

\section*{Acknowledgements}
This work was partially supported by the Beijing Natural Science Foundation (No. L233008).

{
    \small
    \bibliographystyle{ieeenat_fullname}
    \bibliography{main}
}
\appendix


\newcommand{\finished}[1]{{\color{green}{\bf finished}{#1}}}
\newcommand{\TBD}[2]{{\color{red}{\bf TBD}{#2}}}

\newcommand{\qin}[1]{{\textcolor{red}{[Qin: #1]}}}
%




\clearpage
\setcounter{page}{1}
\maketitlesupplementary

\section{Additional Quantitative Results}

\subsection{General-purpose Part Assembly is Beneficial}
We have trained a versatile model (``ours-general'') on all available object categories, which aims to perform part assembly across a variety of categories, and category-specific models (``ours-specific''), where each model is tailored to a specific category.

\noindent\textbf{Influence on Assembly Sequence Length.}
We analyze the impact of multi-task learning on varied assembly lengths in~\cref{tab:multitask_longhorizon}.
Compared to the category-specific models, the multi-task model continues to gain a notable improvement of $+2.77\%$ in long-horizon assembly, and also improve by $+1.9\%$ in short-horizon assembly.

\begin{table}[h]
\centering
\caption{Average results across all categories. The numbers in brackets denote the object number in the test set.}
\scalebox{0.7}{
\begin{tabular}{cc@{}|ccc@{}|cc}
\toprule
&&\multicolumn{2}{c}{$\leq$10 parts (3318)} && \multicolumn{2}{c}{$>$10 parts (1223)}\\
\midrule
\multicolumn{1}{c}{\centering Method}& &\multicolumn{1}{c}{PA} & \multicolumn{1}{c}{SR}&& PA & SR\\
\midrule
ours-specific&& 61.74 & 36.20 && 
53.25 & 7.59\\
ours-multitask&& 63.64 & 37.46 && 
56.02 & 9.31\\
\bottomrule
\end{tabular}}
\label{tab:multitask_longhorizon}
\end{table}

\noindent\textbf{Detailed Results on Daily Object Assembly.}
In ~\cref{supple:comp_1_all_cat}, we conduct comparisons between the versatile model and category-specific models. 
Our versatile model demonstrates significant improvements in categories with both abundant (e.g., tables, storage furniture) and limited (e.g., mugs with 120 samples, bowls with 130 samples) training data.
Given these promising results, we will further explore this direction, such as discovering which categories benefit most from shared assembly knowledge, or how to effectively manage the output categories.

\subsection{More Details and Analysis on Category-specialized Part Assembly}
Unlike the less-explored general-purpose assembly mentioned above, the experiments presented in our main paper follow the traditional benchmark of category-specialized part assembly, which has been commonly adopted by previous works.
In this subsection, we provide additional experimental details.


\noindent\textbf{Additional Implementation Details.}
In our approach, each part is represented by a point cloud, consisting of 1,000 points obtained through Furthest Point Sampling~\cite{eldar1997farthest}.
We ensure these sampled point clouds are zero-centered and aligned with the principal axes determined by Principal Component Analysis (PCA)~\cite{2002pca}.
In our experimental results, the shape chamfer distance (SCD) metric is scaled up by a factor of 1000, while the part accuracy (PA), connectivity accuracy (CA), and success rate (SR) are expressed as percentages.
Moreover, the assembly sequences are constructed based on the relative positions of parts in their original configuration.

\noindent\textbf{Additional evaluation on Per-class Part Assembly. }
To offer a comprehensive comparison of different object categories, we have included per-class evaluation results in~\cref{supple:comp_1_all_cat}.
Our proposed method consistently outperforms previous works in most categories.
It's important to note that the success rates vary among categories, 
We attribute this variation more to the diversity of object structures and fundamental geometries encountered during training, rather than the quantity of training samples.
For example, categories with a large number of training samples, such as faucet (with approximately 460 samples) and lamp (with approximately 1420 samples), do not necessarily yield higher success rates.
Specifically, faucets and lamps demonstrate success rates of only 5.3\% and 17.2\%, respectively, when using our category-specific model.
This is largely due to the complex assembly required by the intricate geometrical composition of their parts.

\begin{table*}
\centering
\caption{Comparisons of per-class results across 18 daily objects, where ``ours-multitask'' and ``ours-specific'' denote the unified model for multi-task assembly and the class-specific models, respectively.
The abbreviations ``Dish'', ``Disp'', ``Ear'', ``Fauc'' and  ``Frid'' denote Dishwasher, Display, Earphone, Faucet and Refrigerator, respectively.
}
\scalebox{0.66}{
\begin{tabular}{ccc|cccccccccccccccccc}
\hline
\multicolumn{1}{c|}{} &
  \multicolumn{1}{c|}{Method} &
  Avg &
  Bag &
  Bed &
  Bottle &
  Bowl &
  Clock &
  Dish &
  Disp &
  Door &
  Ear &
  Fauc &
  Hat &
  Knife &
  Lamp &
  Lap &
  Mug &
  Frid &
  Trash &
  Vase \\ \hline
\multicolumn{1}{c|}{} &
  \multicolumn{1}{c|}{DGL} &
  6.57 &
  23.7 &
  20.2 &
  6.0 &
  20.2 &
  \textbf{3.1} &
  7.0 &
  \textbf{1.4} &
  4.5 &
  16.6 &
  13.8 &
  3.6 &
  \textbf{2.0} &
  \textbf{8.0} &
  \textbf{0.9} &
  5.2 &
  5.8 &
  9.1 &
  2.4 \\
\multicolumn{1}{c|}{} &
  \multicolumn{1}{c|}{RGL} &
  10.93 &
  11.8 &
  9.6 &
  3.5 &
  22.4 &
  9.1 &
  4.5 &
  3.6 &
  3.7 &
  27.2 &
  17.2 &
  9.6 &
  3.4 &
  15.8 &
  17.1 &
  7.8 &
  4.2 &
  4.7 &
  8.7 \\
\multicolumn{1}{c|}{} &
  \multicolumn{1}{c|}{Score} &
  \textbf{6.38} &
  \textbf{1.3} &
  19.8 &
  13.6 &
  \textbf{0.6} &
  7.1 &
  7.8 &
  2.9 &
  9.7 &
  \textbf{13.6} &
  \textbf{9.0} &
  \textbf{1.6} &
  10.4 &
  12.7 &
  12.7 &
  \textbf{1.4} &
  \textbf{2.5} &
  3.6 &
  \textbf{1.2} \\
\multicolumn{1}{c|}{} &
  \multicolumn{1}{c|}{ours-specific} &
  8.61 &
  7.6 &
  \textbf{7.7} &
  \textbf{3.0} &
  23.3 &
  7.7 &
  4.3 &
  3.3 &
  \textbf{3.4} &
  24.5 &
  11.2 &
  9.2 &
  3.3 &
  11.7 &
  5.1 &
  8.5 &
  2.8 &
  4.7 &
  8.9 \\
\multicolumn{1}{c|}{\multirow{-5}{*}{SCD}} &
  \multicolumn{1}{c|}{ours-multitask} &
  8.60 &
  8.5 &
  12.0 &
  3.6 &
  10.7 &
  8.8 &
  \textbf{3.5} &
  4.4 &
  6.5 &
  24.4 &
  12.4 &
  10.8 &
  3.3 &
  11.6 &
  4.0 &
  7.0 &
  2.9 &
  \textbf{2.7} &
  8.7 \\ \hline
\multicolumn{1}{c|}{} &
  \multicolumn{1}{c|}{DGL} &
  36.55 &
  21.7 &
  11.8 &
  80.4 &
  63.0 &
  45.2 &
  11.3 &
  62.9 &
  26.7 &
  27.8 &
  13.1 &
  36.7 &
  41.0 &
  32.8 &
  65.2 &
  13.1 &
  16.5 &
  11.2 &
  23.3 \\
\multicolumn{1}{c|}{} &
  \multicolumn{1}{c|}{RGL} &
  51.33 &
  36.2 &
  13.3 &
  86.1 &
  46.3 &
  52.0 &
  45.2 &
  82.6 &
  \textbf{56.7} &
  41.0 &
  33.7 &
  55.1 &
  82.0 &
  33.3 &
  60.0 &
  \textbf{59.5} &
  48.5 &
  49.4 &
  47.3 \\
\multicolumn{1}{c|}{} &
  \multicolumn{1}{c|}{Score} &
  32.39 &
  \textbf{62.8} &
  1.5 &
  2.2 &
  72.2 &
  11.6 &
  1.8 &
  10.2 &
  10.2 &
  14.0 &
  23.5 &
  58.2 &
  59.8 &
  34.9 &
  62.8 &
  54.8 &
  32.5 &
  20.6 &
  51.4 \\
\multicolumn{1}{c|}{} &
  \multicolumn{1}{c|}{ours-specific} &
  {54.32} &
  52.2 &
  \textbf{28.0} &
  \textbf{87.8} &
  42.6 &
  \textbf{58.4} &
  52.5 &
  \textbf{83.6} &
  47.1 &
  \textbf{52.0} &
  39.7 &
  \textbf{62.2} &
  76.2 &
  34.2 &
  87.2 &
  37.0 &
  52.1 &
  59.4 &
  48.5 \\
\multicolumn{1}{c|}{\multirow{-5}{*}{PA}} &
  \multicolumn{1}{c|}{ours-multitask} &
  \textbf{59.93} &
  46.4 &
  11.0 &
  84.8 &
  \textbf{77.8} &
  56.0 &
  \textbf{62.0} &
  82.2 &
  42.0 &
  50.5 &
  \textbf{42.8} &
  55.1 &
  \textbf{84.0} &
  \textbf{36.9} &
  \textbf{90.2} &
  55.9 &
  \textbf{55.7} &
  \textbf{65.6} &
  \textbf{52.7} \\ \hline
\multicolumn{1}{c|}{} &
  \multicolumn{1}{c|}{DGL} &
  40.01 &
  29.8 &
  21.8 &
  76.5 &
  64.3 &
  \textbf{41.1} &
  14.7 &
  66.3 &
  58.1 &
  38.0 &
  29.7 &
  36.5 &
  42.5 &
  38.7 &
  31.3 &
  9.4 &
  28.7 &
  10.8 &
  27.4 \\
\multicolumn{1}{c|}{} &
  \multicolumn{1}{c|}{RGL} &
  50.52 &
  23.4 &
  12.6 &
  76.5 &
  \textbf{71.4} &
  36.9 &
  27.5 &
  76.0 &
  \textbf{83.8} &
  42.2 &
  47.5 &
  50.0 &
  74.0 &
  \textbf{47.4} &
  71.3 &
  \textbf{11.3} &
  37.3 &
  20.7 &
  32.21 \\
\multicolumn{1}{c|}{} &
  \multicolumn{1}{c|}{Score} &
  32.16 &
  16.2 &
  19.4 &
  24.8 &
  50 &
  26.2 &
  10.9 &
  19.8 &
  19.6 &
  24.1 &
  37.8 &
  \textbf{55.8} &
  51.4 &
  41.4 &
  16.2 &
  \textbf{11.3} &
  32.5 &
  9.8 &
  \textbf{38.5} \\
\multicolumn{1}{c|}{} &
  \multicolumn{1}{c|}{ours-specific} &
  {51.60} &
  \textbf{42.6} &
  \textbf{23.2} &
  \textbf{77.1} &
  57.1 &
  37.6 &
  29.4 &
  \textbf{80.2} &
  70.4 &
  \textbf{55.1} &
  44.2 &
  51.9 &
  \textbf{77.3} &
  46.9 &
  \textbf{95} &
  1.9 &
  50.4 &
  27.6 &
  26.4 \\
\multicolumn{1}{c|}{\multirow{-5}{*}{CA}} &
  \multicolumn{1}{c|}{ours-multitask} &
  \textbf{52.86} &
  \textbf{42.6} &
  15.0 &
  71.2 &
  42.8 &
  \textbf{41.1} &
  \textbf{56.8} &
  77.8 &
  63.7 &
  \textbf{55.1} &
  \textbf{53.9} &
  44.2 &
  76.2 &
  47.2 &
  91.2 &
  9.4 &
  \textbf{55.6} &
  \textbf{33.0} &
  31.2 \\ \hline
\multicolumn{1}{c|}{} &
  \multicolumn{1}{c|}{DGL} &
  18.73 &
  13.8 &
  0 &
  69.0 &
  53.8 &
  23.9 &
  0 &
  37.7 &
  3.9 &
  0 &
  0 &
  13.3 &
  7.8 &
  14.5 &
  47.6 &
  0 &
  0 &
  0 &
  11.6 \\
\multicolumn{1}{c|}{} &
  \multicolumn{1}{c|}{RGL} &
  32.0 &
  17.2 &
  0 &
  75.0 &
  25.6 &
  47.7 &
  13.7 &
  67.5 &
  \textbf{47.1} &
  1.9 &
  1.5 &
  24.4 &
  59.7 &
  14.0 &
  41.5 &
  \textbf{31.4} &
  12.9 &
  26.3 &
  40.0 \\
\multicolumn{1}{c|}{} &
  \multicolumn{1}{c|}{Score} &
  15.05 &
  \textbf{34.5} &
  0 &
  0 &
  66.7 &
  3.4 &
  0 &
  8.4 &
  11.8 &
  0 &
  1.5 &
  26.7 &
  19.5 &
  10.1 &
  12.2 &
  25.7 &
  3.2 &
  2.6 &
  43.6 \\
\multicolumn{1}{c|}{} &
  \multicolumn{1}{c|}{ours-specific} &
  {37.12} &
  31.0 &
  \textbf{6.3} &
  \textbf{78.6} &
  25.6 &
  \textbf{53.4} &
  15.7 &
  \textbf{69.7} &
  45.1 &
  \textbf{17.0} &
  5.3 &
  \textbf{37.8} &
  50.6 &
  \textbf{17.2} &
  78.0 &
  5.7 &
  16.2 &
  \textbf{42.1} &
  42.2 \\
\multicolumn{1}{c|}{\multirow{-5}{*}{SR}} &
  \multicolumn{1}{c|}{ours-multitask} &
  \textbf{39.06} &
  24.1 &
  0 &
  75.0 &
  \textbf{71.8} &
  46.6 &
  \textbf{25.5} &
  68.0 &
  33.3 &
  \textbf{17.0} &
  \textbf{10.6} &
  26.7 &
  \textbf{62.3} &
  15.7 &
  \textbf{82.9} &
  \textbf{31.4} &
  \textbf{19.4} &
  39.5 &
  \textbf{47.8} \\ \hline
\multicolumn{3}{c|}{Test Number} &
  29 &
  16 &
  84 &
  39 &
  88 &
  51 &
  191 &
  51 &
  53 &
  132 &
  45 &
  77 &
  407 &
  82 &
  35 &
  31 &
  38 &
  232 \\ \hline
\end{tabular}
}
\label{supple:comp_1_all_cat}
\end{table*}

\section{Additional Qualitative Examples}
\begin{figure*}[t]
  \centering
  \vspace{-5pt}	
\includegraphics[width=0.9\linewidth]{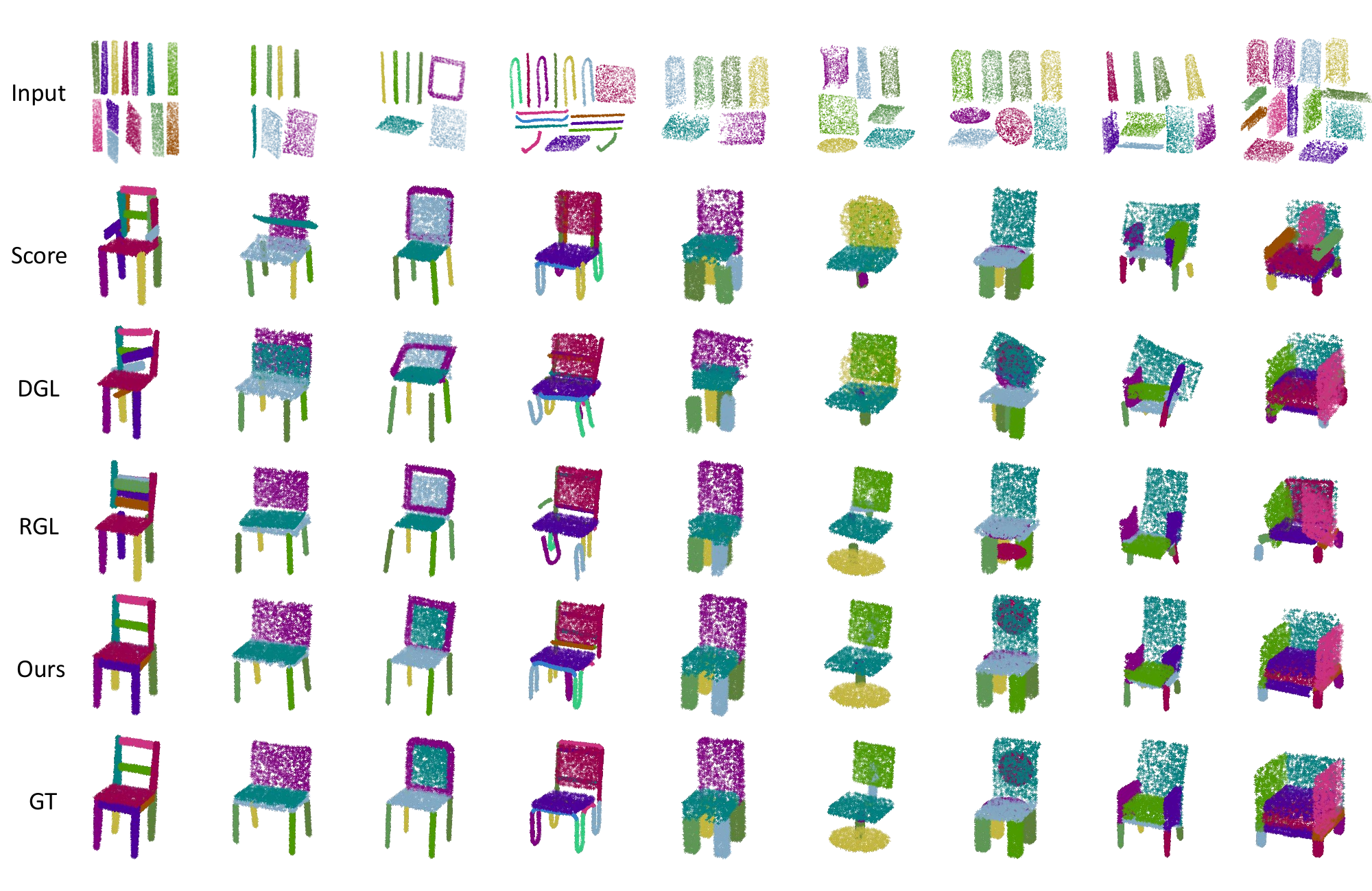}  
  \caption{Qualitative results and comparisons on the chair assembly task. Distinct colors within a single shape denote various parts of the chair, whereas consistent coloring in a row signifies identical parts. Our SPAFormer is able to identify and adhere to appropriate assembly patterns to ensure accurate assembly of structured objects.}
  \label{fig:vis_chairs}
  \vspace{-5pt}
\end{figure*}

\begin{figure*}[t]
  \centering
	\includegraphics[width=0.75\linewidth]{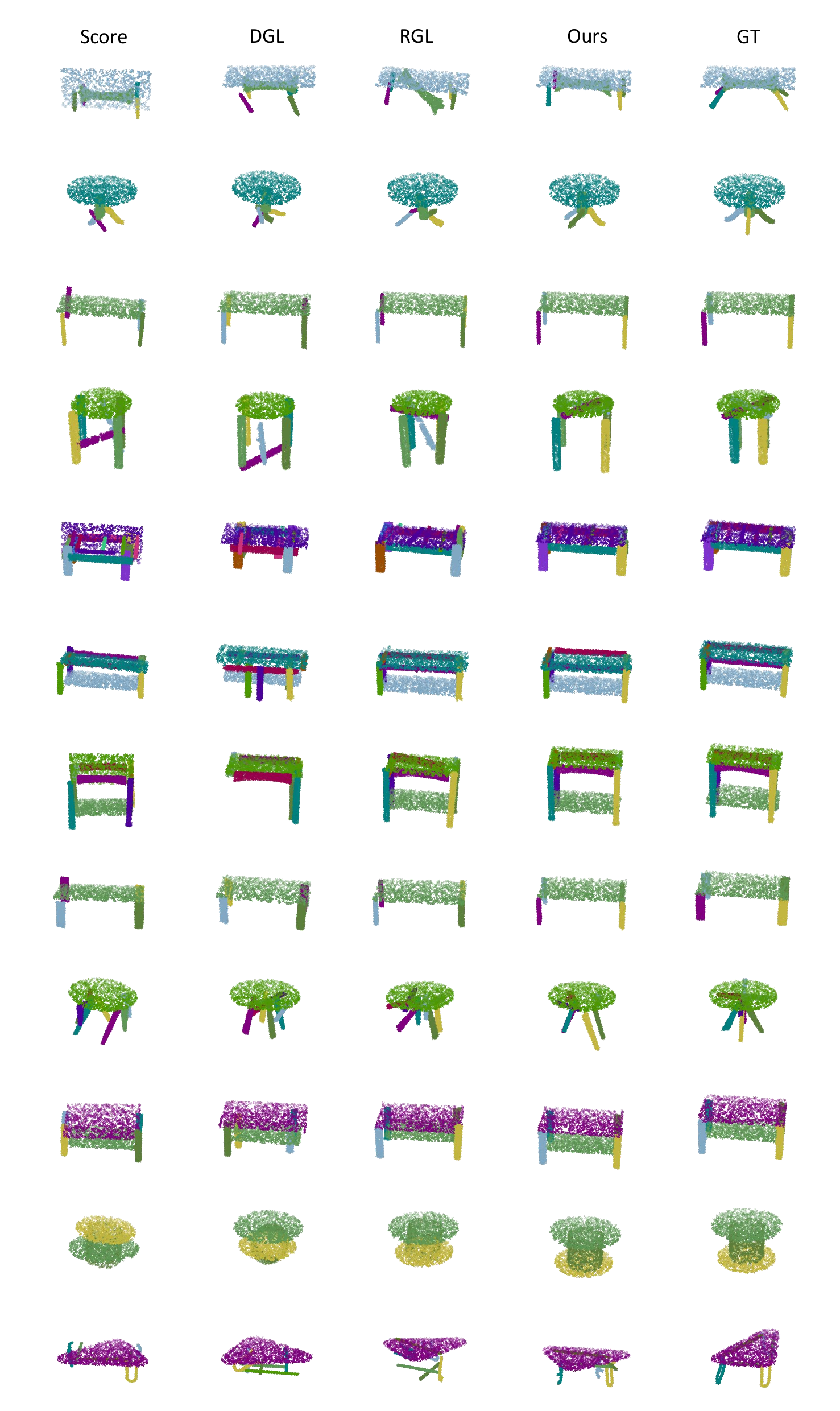}  

  \caption{Qualitative results and comparisons on tables.}
  \label{fig:vis_table}
\vspace{-5pt}	
\end{figure*}

\begin{figure*}[t]
  \centering
	\includegraphics[width=0.75\linewidth]{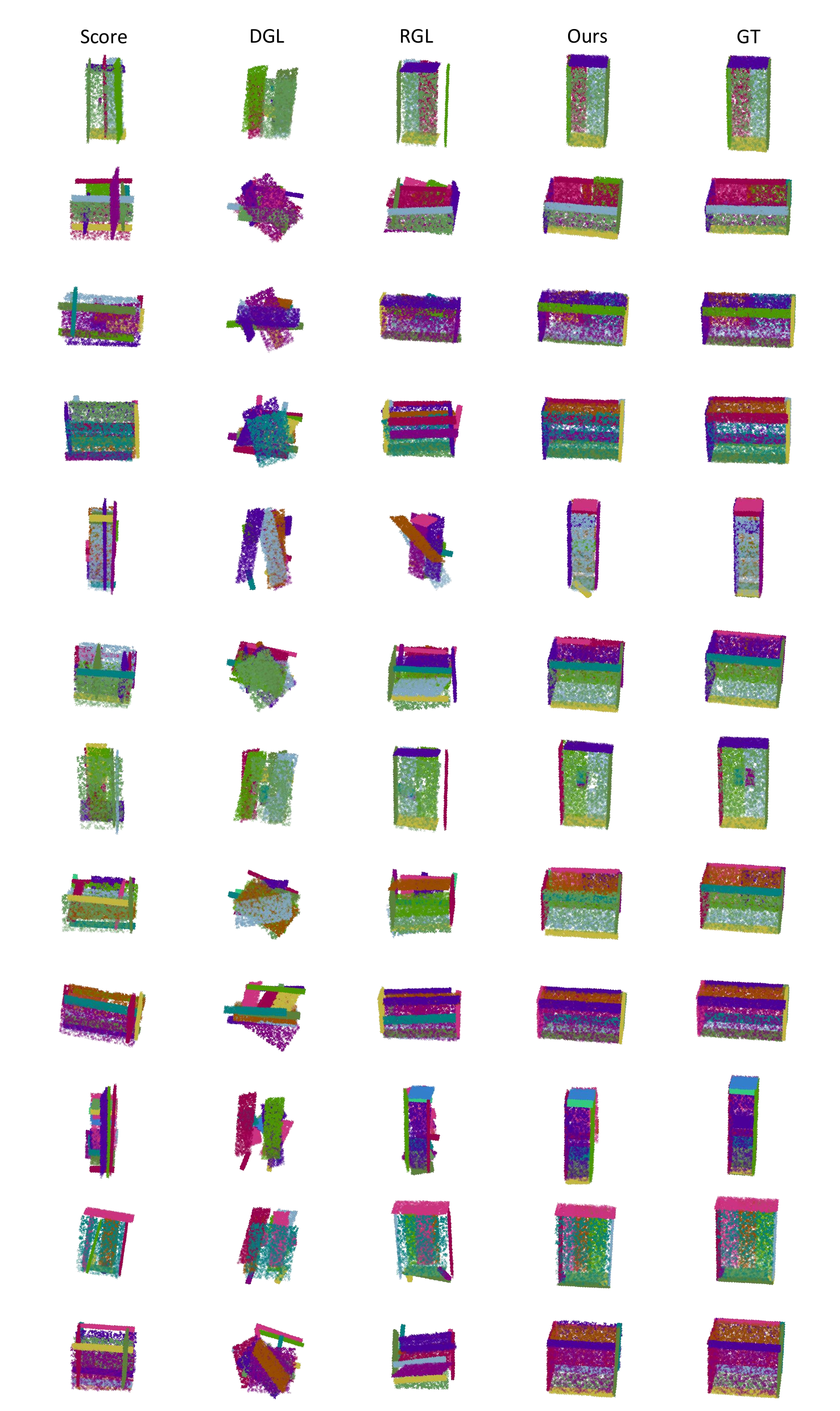}  

  \caption{Qualitative results and comparisons on storage furniture.}
  \label{fig:vis_storage}
\vspace{-5pt}	
\end{figure*}

Additional visualizations, including both successful and unsuccessful cases, are presented in~\cref{fig:vis_chairs,fig:vis_table,fig:vis_storage}.
Our approach not only excels in effectively assembling structural tables that have small connection areas but also shows significant improvement in assembling larger parts of storage furniture, which typically have more extensive connection areas.

\section{Limitations and Directions}

First, acquiring the assembly sequence is sometimes unfeasible, especially when predicted by assembly sequence planning algorithms.
On the other hand, while assembly sequences can be derived from various sources such as instructional manuals, the goal for a truly autonomous agent is to independently determine assembly sequences and navigate the assembly process on its own, mirroring human capabilities.
Therefore, combining the task of assembly sequence planning and 3D part assembly would significantly benefit the object assembly problem.

Second, current methods for 3D-PA primarily focus on point cloud processing without motion planning in the real world, which limits the application in robotics.
Working towards this path and deploying the models in the real world can significantly expand the scope of this problem.

\clearpage


\end{document}